# Unidirectional-Road-Network-Based Global Path Planning for Cleaning Robots in Semi-Structured Environments

Yong Li[1,2,*], *Member, IEEE,* and Hui Cheng[2]

*Abstract—* Practical global path planning is critical for commercializing cleaning robots working in semi-structured environments. In the literature, global path planning methods for free space usually focus on path length and neglect the traffic rule constraints of the environments, which leads to high-frequency re-planning and increases collision risks. In contrast, those for structured environments are developed mainly by strictly complying with the road network representing the traffic rule constraints, which may result in an overlong path that hinders the overall navigation efficiency. This article proposes a general and systematic approach to improve global path planning performance in semi-structured environments. A unidirectional road network is built to represent the traffic constraints in semi-structured environments and a hybrid strategy is proposed to achieve a guaranteed planning result. Cutting across the road at the starting and the goal points are allowed to achieve a shorter path. Especially, a two-layer potential map is proposed to achieve a guaranteed performance when the starting and the goal points are in complex intersections. Comparative experiments are carried out to validate the effectiveness of the proposed method. Quantitative experimental results show that, compared with the state-of-art, the proposed method guarantees a much better balance between path length and the consistency with the road network.

## I. Introduction

### A. Motivation

As a typical service robot, the cleaning robot is used to clean the solid and liquid wastes on the ground [1]. Cleaning robots in unstructured environments have been widely used, such as the household-sweeping robot [2], while the ones working in garages, construction zones, around shopping centers, and other semi-structured environments are still at an early stage [3][4]. There are still many problems to be solved, one of them is finding a practical global path.

In addition to accomplishing the cleaning task and ensuring their own safety, robots working in semi-structured environments need to interact with vehicles, non-vehicles, pedestrians, and even other kinds of robots and autonomous vehicles. So as not to cause confusion or trouble to human drivers and other agents, they should also try to comply with the traffic rules. A practical approach is building a road network that accounts for the traffic rules so that we can plan a global path with it [5]. In the fields of autonomous driving and automated valet parking, road-network-based global path planning is already a common practice [4][6][7]. The basic flow is: (1) find the nodes pair $\{p_{ms}, p_{mg}\}$ closest to the starting point $p_s$ and the goal point $p_g$ in the road network respectively ( $p_{ms}$ and $p_{mg}$ should also satisfy the angle constraint); (2) use graph search algorithm to find the path connecting $p_{ms}$ and $p_{mg}$; (3) adjust path points density and smooth the path. However, such practices may not be the best solution for robots in semi-structured environments. As Fig.1 shows, strictly following the traffic rules results in overlong paths in some situations, which results in low working efficiency for cleaning robots.

Balancing path length and the consistency with the road network is critical for commercializing cleaning robots working in semi-structured environments and is the main focus of this article.

### B. Related Work

Common algorithms for global path planning can be classified into four types[8]: graph-search-based planners, sampling-based methods, interpolating-curve-based and optimization-based ones. Detailed analysis and comparisons can be seen in review articles [8] and [9].

Currently, most of the research on global path planning is for unstructured and structured environments [7]. For the former, global path planning is viewed as a free-space-path-finding problem. Methods in the literature are proposed mainly to shorten plan length and improve search efficiency with/without kinematics constraints [10] [11][12]. Such methods are not suitable for semi-structured environments as they do not take traffic rules and the respect for human drivers into account, which brings enormous safety risks to both the robot itself and other traffic participants.

The global path planning methods in structured environments are usually developed based on the traffic rule constraints described by unidirectional road networks [7][13] [14]. The planning results are required to strictly comply with the traffic rules such as lane following, lane changing, merging, pulling over, and so on [15]. The robot is not allowed to cut across the road or drive reversely on highways. As shown in Fig. 1, strictly complying with the traffic rules negatively impacts the robot's work efficiency.

Some research has been carried out to meet the set-out requirements in semi-structured environments. Tsiakas uses a sparse road network described by OSM to guide the global path planning process, and pathfinding is achieved by the A* algorithm [16]. Klaudt combined the road network with a semantic and metric map to realize parking path planning in garages using a state-based planner [4]. The above two research work is based on the bidirectional road network. The planned global path points are usually distributed along the road center, which does not comply with the right-hand traffic (or left-hand traffic) rule, increasing the re-planning frequency and the risk of collision. Dolkov combined the free spa-ce hybrid A* algorithm with the road network to improve the path quality in semi-structured environments [17][18]. In the nodes expansion step of the A* search, the deviation from the road network is penalized, and the road network nodes also provide a good set of macro-actions. The algorithm ensures that the planned path points are close to the road network but can not guarantee they align with the road network in direction.

[1]Guangzhou Shiyuan Electronic Technology Co., Ltd., Guangzhou 510300, China.
[2]School of Data and Computer Science, Sun Yat-sen University, Guangzhou 510006, China.
[*]Corresponding author: Yong Li (e-mail: liyong2018@zju.edu.cn).

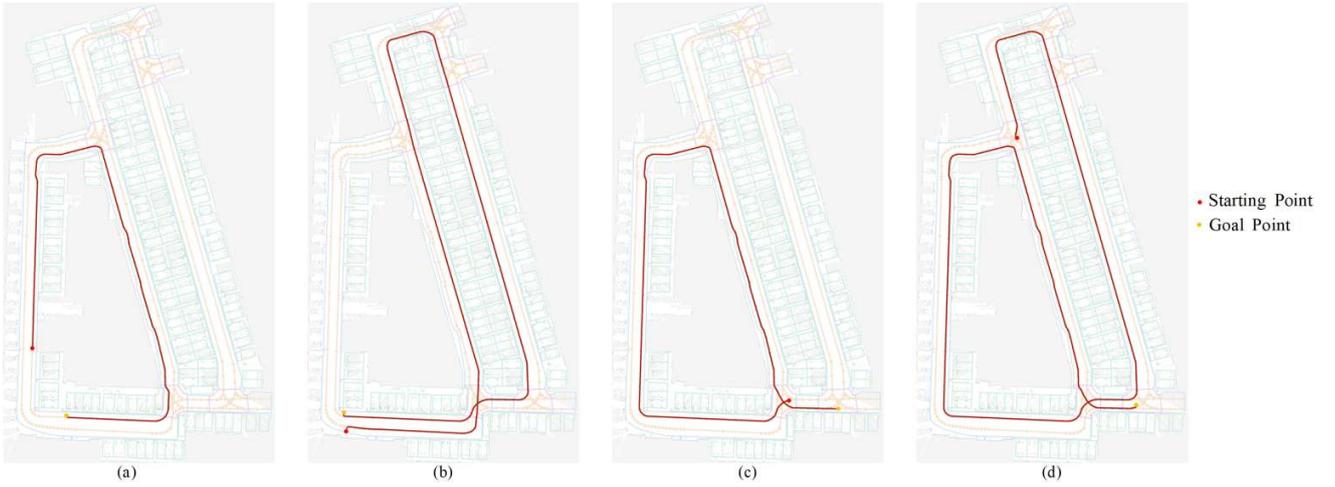

Fig. 1. Global path planning results when strictly complying with the road network that accounts for the traffic rules: (a) the starting and goal points are in the same lane and the latter is behind the former; (b) The starting and goal points are close but they are in the reverse lanes; (c) The starting and goal points are in different intersections and they are close to each other; (d) The starting and goal points are in different intersections and they are far away from each other.

*C. Contributions*

This article proposes a practical global path planning algorithm for commercial cleaning robots working in semi-structured environments. The main contributions are:

(1) A general and systematic global path planning algorithm based on a unidirectional road network and two-layer potential map is proposed, which makes a better balance between path length and the consistency with the road network (both in distance and direction).

(2) A scenario-based strategy is adopted to meet the set-out requirements in semi-structured environments. The robot is allowed to cut across the road at the starting and goal points, which ensures the finding of a shorter path. Besides, the road network constraints in complex intersections are described with a two-layer potential map.

(3) Comparative experiments are carried out and quantitative performance indexes are introduced to verify the superiority of the proposed method.

The rest of the article is organized as follows: Sec. II is about problem description, followed by Sec. III presenting the methodology. Then comparative and field experimental results are described in Sec. IV. Conclusions are drawn in Sec. V.

## II. PROBLEM DESCRIPTION

Without loss of generality, the descriptions in the rest of the article are based on a garage with the traffic rule of right-hand driving, and the research results can be easily extended to other semi-structured environments. Fig. 2 is a typical semantic map for the garage-cleaning robot. The passable areas include the passages (chocolate), the intersections (dark violet) and the parking areas (light green). The cleaning robot needs to complete the full coverage cleaning task in passable areas (full coverage path planning in the semi-structured environment will be carried out in our following research and this study only focuses on global path planning when the robot travels between different clean areas). Actually, appropriately cutting across the road is acceptable [19] to some extent for the following two reasons. First, when the robot executes the full coverage cleaning task, it is unavoidable that the robot cuts across the road; It is also reasonable to allow such behavior in global path planning. Second, the cleaning robot is much smaller than the vehicles. It is usually driven by differential wheels and thus has a much smaller turning radius than Ackerman-type robots. To balance safety and movement flexibility, the shortcut is only allowed at the starting and goal points in this study.

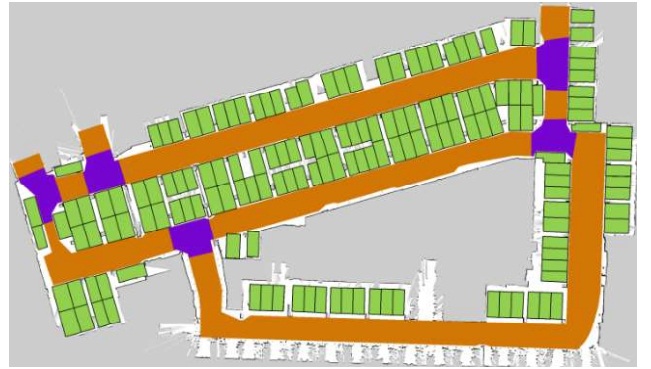

Fig. 2. The passable areas for a garage-cleaning robot with passages in chocolate, intersections in dark violet and parking areas in light green.

Besides, we should map the starting and goal points to the road network to find a valid path. However, finding a suitable mapping is challenging when the starting and goal points are in complex intersections even with the allowance of the shortcut. Noticeably, the road network has two kinds of restraints to the global path planning process: distance constraint and direction constraint. Considering the complexity of intersections and the fact that intersections usually have small areas, it is reasonable and practical to only consider the distance constraint in intersections.

For a road network represented as $G = \{V, E\}$ with $V$ nodes and $E$ edges, our goal is to minimize the planning cost $J(S)$ as follows:

$$J(S) = \min_{S}\{f_l(S) + \sum_{i=0}^{n} f_d(s_i, G) + \sum_{i=0}^{n} f_\theta(s_i, G)\} \quad (1)$$

where $S = \{s_i, i = 1,2,\ldots,n\}$ is the global path with $s_i = \{x_i, y_i, \theta_i\}$ the $i$th global path point, $f_l(S)$ the path length cost, $f_d(s_i, G)$ and $f_\theta(s_i, G)$ the cost of the distance and angle between $s_i$ and $G$, respectively. This article tries to minimize $J(S)$ with hybrid strategies rather than giving an analytic solution.

## III. METHODOLOGY

### A. Unidirectional road network

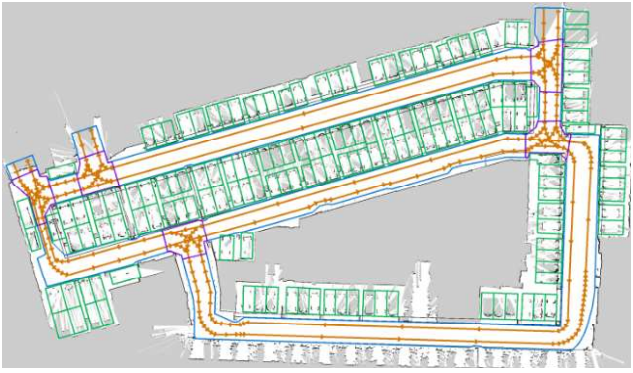

Fig. 3. A typical semantic map for the garage-cleaning robot with the unidirectional road network in chocolate, intersection areas in dark violet and parking areas in light green. The arrows of the lanes indicate their direction.

The road network can be designed according to the actual situation of the environment with considering the set-out requirements. For example, the unidirectional road network for the garage of Guangzhou Shiyuan Electronic Technology Co., Ltd. is designed as Fig. 3 shows. The road network is composed of unidirectional lanes $L = \{L_i, i = 1,2,\ldots,n\}$ and each lane $L_i$ has two nodes $p_i = \{p_{is}, p_{ie}\}$ (fewer nodes improve the speed of road network construction and pathfinding process). The predecessor lanes $L_i^{from}$ and the successor lanes $L_i^{to}$ of $L_i$ can be expressed as:

$$L_i^{from} = \{L_j \in L : |\theta_{p_{je}} - \theta_{p_{is}}| < \alpha_{min} \,\&\, \|p_{je} - p_{is}\| < d_{min}\}$$

$$L_i^{to} = \{L_j \in L : |\theta_{p_{js}} - \theta_{p_{ie}}| < \alpha_{min} \,\&\, \|p_{js} - p_{ie}\| < d_{min}\}$$
(2)

where $\theta_*$ is the angle of node $*$ ($*= p_{je}, p_{is}, p_{js}$ and $p_{ie}$), $\alpha_{min}$ the angle threshold and $d_{min}$ the distance threshold. Besides, the procedure for finding the reverse lanes $L_i^{reverse}$ of $L_i$ can be seen in Algorithm 1. For every node in $L_i$, the algorithm tries to find its closest lane $L_j$ in $L$ with $L_j$ no a predecessor/successor lane nor an intersection lane of $L_i$. The founded lane and $L_i$ are reverse lanes of each other. The nodes of $L$ are also the nodes $V$ of road network $G$. the edges $V$ of $G$ are represented as unidirectional edges. The predecessor node $p_{ie}^{from}$ of $p_{ie}$ is $p_{is}$ while the successor node $p_{is}^{to}$ of $p_{is}$ is $p_{ie}$. The predecessor nodes $p_{is}^{from}$ of $p_{is}$ and successor nodes $p_{ie}^{to}$ of $p_{ie}$ are respectively represented as:

$$p_{is}^{from} = \{p_{je} : p_{je} \in L_j, L_j \in L_i^{from}\}$$

$$p_{ie}^{to} = \{p_{js} : p_{js} \in L_j, L_j \in L_i^{to}\} \quad (3)$$

---
**Algorithm 1: Procedure in Finding Reverse Lane**

1   **for** all $L_i \in L$ **do**
2     **for** all $p_i \in L_i$ **do**
3       $d_{min} \leftarrow +\infty$
4       $id \leftarrow -1$
5       **for** all $L_j \in L$ **do**
6         **if** $L_j.id \mathrel{!}= L_i.id$ **then**
7           **if** $L_j \notin L_i^{from}$ **then**
8             **if** $L_j \notin L_i^{to}$ **then**
9               **if** $L_j$ does not intersects with $L_i$ **then**
10                $\mathbf{P} \leftarrow \overrightarrow{p_{js}p_{je}}$
11                $\mathbf{P_0} \leftarrow \overrightarrow{p_{js}p_i}$
12                $\mathbf{P_1} \leftarrow \overrightarrow{p_{je}p_i}$
13                $d \leftarrow \dfrac{\mathbf{P_0} \times \mathbf{P}}{\|\mathbf{P}\|}$
14                $\Delta\theta \leftarrow \theta_{p_i} - \theta_{\mathbf{P}}$
15                **if** $\Delta\theta > \theta_{threhood}$ **then**
16                   **if** ($\mathbf{P}\mathbf{P_0} > 0$ && $\mathbf{P}\mathbf{P_1} < 0$) **then**
17                     **if** $d < d_{threhood}$ **then**
18                       **if** $d < d_{min}$ **then**
19                         $d_{min} \leftarrow d$
20                         $id \leftarrow j$

21     **if** $id > 0$ **then**
22       $L_i^{reverse} \leftarrow L_i^{reverse} \cup L_{id}$
23       $L_{id}^{reverse} \leftarrow L_{id}^{reverse} \cup L_i$

---

Based on the above nodes and edges, the unidirectional road network $G = \{V, E\}$ can be built.

### B. Search strategy

**Case1: path planning with starting and goal points in passages or parking areas.** When the starting and goal points are in passages or parking areas, the complete planning procedure is shown in Algorithm 2. In this study, the starting and goal points are mapped not only to the closet lane but also to the reverse lanes of their closet lane. For the planned path, the mapping rules are designed to ensure no sharp turns around the matched starting points or the matched goal ones.

*1) Starting point mapping*: For starting point $p_s$, donate its closest lane as $L_c$ and $L_c$'s reverse lane is $L_c^{reverse}$. The set of $L_c$ and $L_c^{reverse}$ is represented as $\Psi = \{L_c \cup L_c^{reverse}\}$. For $\forall \Psi_k \in \Psi, k = 1,2,\ldots,m$ with $\Psi_k$ composed of $p_{ks}$ and $p_{ke}$, the relative spatial relationship between $p_s$ and $\Psi_k$ can be seen in Fig. 4 with $p_{s1}, p_{s2}$ and $p_{s3}$ three possible locations of $p_s$. For easy of notation, let $\mathbf{P_0} = \overrightarrow{p_{ks}p_s}$, $\mathbf{P_1} = \overrightarrow{p_{ke}p_s}$, $\mathbf{P} = \overrightarrow{p_{ks}p_{ke}}$. The matching point $p_k^{match}$ can be expressed as follows:

- if $\mathbf{P_0}\mathbf{P} < 0$ (corresponding to $p_{s1}$ in Fig.4(a)), $p_k^{match} = p_{ks}$.

- if $\mathbf{P_0}\mathbf{P} > 0 \,\&\, \mathbf{P}\mathbf{P_1} < 0$ (corresponding to $p_{s2}$ in Fig. 4(a)), $p_k^{match} = p_{ks} + \mathbf{P_0}\mathbf{P}/\|\mathbf{P}\|^2$.

- if $\mathbf{P}\mathbf{P_1} > 0$ (corresponding to $p_{s3}$ in Fig. 4(a)), $p_k^{match} = \{p_{us} \in L_u, L_u \in \Psi_k^{to}\}$.

Then the set of the matched starting points is represented as $\boldsymbol{P}_{start}^{match} = \{p_k^{match}, k = 1, 2, \ldots, m\}$.

*2) Goal point mapping:* Similarly, as shown in Fig.4(b), the matching point $p_k^{match}$ for the goal point $p_g$ can be represented as follows:

- if $\boldsymbol{P_0 P} < 0$ (corresponding to $p_{e1}$ in Fig.4(b)), $p_k^{match} = \{p_{vs} \in L_v, L_v \in \Psi_k^{from}\}$.

- if $\boldsymbol{P_0 P} > 0 \,\&\, \boldsymbol{P P_1} < 0$ (corresponding to $p_{e2}$ in Fig. 4(b)), $p_k^{match} = p_{ks} + \boldsymbol{P_0 P P}/\|\boldsymbol{P}\|^2$.

- if $\boldsymbol{P P_1} > 0$ (corresponding to $p_{e3}$ in Fig. 4(b)), $p_k^{match} = p_{ke}$.

Then the set of the matched goal points is represented as $\boldsymbol{P}_{goal}^{match}$. $\boldsymbol{P}_{start}^{match}$ and $\boldsymbol{P}_{goal}^{match}$ serve as the start and goal candidates in the path searching process.

---
**Algorithm 2: Procedure in Case 1**

**input**: Starting point $p_s$, Goal point $p_g$, Road network $G$
**output**: GlobalPath $\chi(p_s, p_g)$

1  $\boldsymbol{P}_{start}^{match} \leftarrow$ FindMatchedStartPoints($p_s, G$)
2  $\boldsymbol{P}_{goal}^{match} \leftarrow$ FindMatchedGoalPoints($p_g, G$)
3  $MaxCost \leftarrow 2 \times G.Nodes.size()$
4  $MinLength \leftarrow +\infty$
5  $SearchCost \leftarrow 0$
6  **for** all $p_s^{match} \in \boldsymbol{P}_{start}^{match}$ **do**
7    **for** all $p_g^{match} \in \boldsymbol{P}_{goal}^{match}$ **do**
8      $bSuccess \leftarrow$ BuildSearchTree($p_s^{match}, p_g^{match}, MaxCost, G, SearchCost$)
9      **if** $bSuccess$ **then**
10       $\{Path, PathLength\} \leftarrow$ getPath()
11       $\beth \leftarrow Path \cup p_s \cup p_g$
12       $d_s \leftarrow Distance(p_s, p_s^{match})$
13       $d_g \leftarrow Distance(p_g, p_g^{match})$
14       $TotalPathLength \leftarrow PathLength + d_s + d_g$
15       **if** $TotalPathLength < MinLength$ **then**
16         $MaxCost \leftarrow 2 \times SearchCost$
17         $\chi(p_s, p_g) \leftarrow \beth$

---

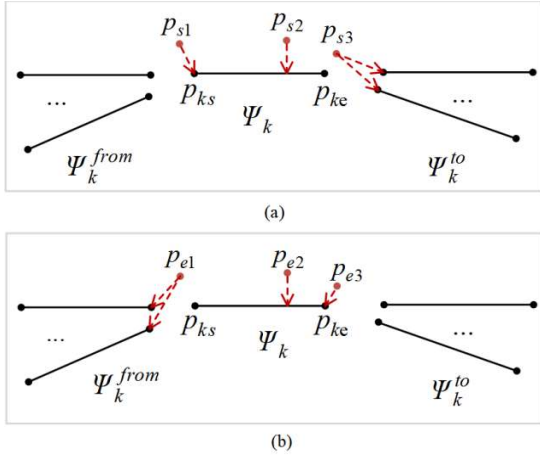

Fig. 4. (a) starting point mapping: $p_{s1}, p_{s2}$ and $p_{s3}$ represent three probable positions of starting point $p_s$, $\Psi_k$ the closest lane to $p_s$, $\Psi_k^{from}$ and $\Psi_k^{to}$ the predecessor and successor lane of $\Psi_k$, respectively. The ends of the arrows represent the mapping results; (b) goal point mapping: $p_{e1}, p_{e2}, p_{e3}$ are three probable positions of goal point $p_g$, $\Psi_k$ the closest lane to $p_g$, $\Psi_k^{from}$ and $\Psi_k^{to}$ the predecessor and successor lane of $\Psi_k$, respectively. The ends of the arrows represent the mapping results.

*3) Path searching:* For every $p_s^{match}$ in $\boldsymbol{P}_{start}^{match}$ and $p_g^{match}$ in $\boldsymbol{P}_{goal}^{match}$, the Dijkstra algorithm is adopted to find a valid path. Appending the starting point $p_s$ and the goal point $p_g$ to the front and back of the Dijkstra path, respectively, then we can get the total path $\beth$ that connects $p_s$ and $p_g$. Looping through $\boldsymbol{P}_{start}^{match}$ and $\boldsymbol{P}_{goal}^{match}$ then we can get the shorted path $\chi$. A branch-and-bound method is adopted in the cycles to improve the computation efficiency: If the length of current $\beth$ is shorter than the shortest path $\chi$ ever, the total search cost of the Dijkstra process in this cycle will be used as the upper bound cost for Dijkstra in the following cycles.

---
**Algorithm 3: Procedure in Case 2**

**input**: StartPose $p_s$, GoalPose $p_g$, Road network $G$, Multi-level map $\Xi$
**output**: GlobalPath $\chi(p_s, p_g)$

1  $\boldsymbol{P}_{start}^{inters} \leftarrow$ FindIntersectedPoints($p_s, G$)
2  $\boldsymbol{P}_{goal}^{inters} \leftarrow$ FindIntersectedPoints($p_g, G$)
3  $MaxCost \leftarrow 2 \times G.Nodes.size()$
4  $MinLength \leftarrow +\infty$
5  $StartTempPoint\ p_{st} \leftarrow p_s$
6  $GoalTempPoint\ p_{gt} \leftarrow p_g$
7  $PathLength \leftarrow 0$
8  **for** all $p_s^{inters} \in \boldsymbol{P}_{start}^{inters}$ **do**
9    **for** all $p_g^{inters} \in \boldsymbol{P}_{goal}^{inters}$ **do**
10     $bSuccess \leftarrow$ BuildSearchTree($p_s^{inters}, p_g^{inters}, MaxCost, G, SearchCost$)
    **if** $bSuccess$ **then**
11       $\{Path, PathLength\} \leftarrow$ getPath()
12       **if** $PathLength < MinLength$ **then**
13         $MaxCost \leftarrow 2 \times SearchCost$
14         $p_{st} \leftarrow p_s^{inters}$
15         $p_{gt} \leftarrow p_s^{inters}$
16         $\chi(p_{st}, p_{gt}) \leftarrow$ Path

17 $\chi(p_s, p_{st}) \leftarrow$ SearchWithDijkstra($\Xi, p_s, p_{st}$)
18 $\chi(p_{gt}, p_g) \leftarrow$ SearchWithDijkstra($\Xi, p_{gt}, p_g$)
19 $\chi(p_s, p_g) \leftarrow \chi(p_s, p_{st}) \cup \chi(p_{st}, p_{gt}) \cup \chi(p_{gt}, p_g)$

---

**Case2: path planning with starting and goal points in intersections.** The procedure in Case 1 can not guarantee the quality of the generated path in Case 2 due to the complexity of intersection areas. In this article, a hybrid strategy is proposed: the sub-paths inside the intersections are obtained based on a two-layer potential map, while those outside the intersections are gained with a strategy similar to that in Case 1. Detailed procedures can be seen in Algorithm 3.

*1) path planning outside the intersections:* For $p_s$ and $p_g$, without loss of generality, assume they locate in intersections $S_m$ and $S_n$ ($m \neq n$), respectively. The set of intersected points of $S_m/S_n$ and $G$ is represented as $\boldsymbol{P}_{start}^{inters}/\boldsymbol{P}_{goal}^{inters}$. Similar to the procedure in Case 1, the shortest path $\chi(p_{st}, p_{gt})$ can be obtained with starting point candidates $\boldsymbol{P}_{start}^{inters}$ and goal point candidates $\boldsymbol{P}_{goal}^{inters}$. The front and back points of $\chi(p_{st}, p_{gt})$ are represented as $p_{st}$ and $p_{gt}$, respectively.

*2) path planning within the intersections:* In this step, sub-paths $\chi(p_s, p_{st})$ and $\chi(p_{gt}, p_g)$ are obtained so that we can get the complete path $\chi(p_s, p_g)$ from $p_s$ to $p_g$. A two-layer potential map is proposed to represent the distance constraints of the road network. As can be seen in Fig.5, the first layer is a traditional static map $m_{static}$ with the obstacles inflated. The second layer $m_{semantic}$ is built based on the semantic information in Fig.3. A Gaussian potential field is generated within the passable area with the lanes reference (similar approaches can be seen in [20] and [21]). For every point $x_i$ in the passable area $U_p$, if the distance between $x_i$ a-

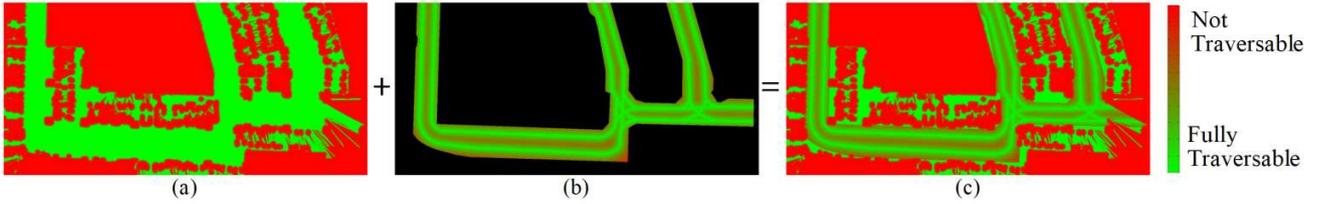

Fig. 5. Part of the two-layer potential map used for path planning when the starting and goal points are in intersections: (a) static metric map with obstacles inflated; (b) road network Gaussian likelihood potential maps; (c) the combined map.

nd the road network $G$ is $d_{x_i}$, its potential value $p(x_i)$ can be computed as:

$$p(x_i) = p_0[1 - exp(-d_{x_i}^2/2\sigma^2)], x_i \in U_p \quad (4)$$

where $p_0$ is the maximal potential, $\sigma$ the standard deviation. The potential value for points on the road boundary is $p_0$ and that for points outside passable areas is 0. Combining $m_{static}$ and $m_{semantic}$ with the adoption of a larger potential value in each grid cell, we can get the map $m$ used for path planning. A traditional Dijkstra planner is then adopted to get $\chi(p_s, p_{st})$ and $\chi(p_{gt}, p_g)$.

It can be seen from the above procedure in Case 2 that a piece-wise planning strategy is adopted: the starting point $p_{st}$ and the goal point $p_{gt}$ for the planning within the intersections are the outcomes of the planning outside the intersections; we end up with a path that may not be optimal. Considering the computation burden and the fact that the intersections often have a small area, the above sub-optimal strategy is acceptable.

It should be noted that this article only describes the path planning strategy in two typical situations, where both the starting and goal points are in the passage/parking area or in different intersections. There are much more combinations such as: the starting point in passage/parking area while the goal point in intersections, the goal point in passage/parking area while the starting point in intersections, or the starting and goal points in the same intersection. Due to space limitations, we will not elaborate on them case by case. Noticeably, they can be easily handled with the basic ideas in Case1&2.

## IV. EXPERIMENTS

### A. Comparative experiments

To verify the effectiveness of the proposed method, comparative experiments are carried out. The computing unit used in the experiments is an industrial computer with CPU i7-10700@2.9Hz×16 and RAM of 16 GB. The size of the map used in the experiment is 75$m$ * 128$m$. The following four methods are compared:

***Hybrid-A\*-in-SS***: Hybrid A\* in semi-structured environments [18], is considered the state-of-the-art. The cost coefficient $C_G$ (the deviation from the road network) is set to be 1.

***Dijkstra***: The widely used Dijkstra algorithm in free space. The map resolution is 0.05m.

***Dijkstra-in-SS***: Dijkstra in semi-structured environments. In the nodes-expansion step of Dijkstra, the cost representing the distance between the current node and the road network is considered.

***Ours***: the method proposed in this article.

To quantify the performance of different planners, the following performance indexes are used:

- $t = (1/100) \sum_{i=0}^{100} t_i$, the average planning time for 100 consecutive planning cycles, is used to evaluate the computation efficiency;

- $l$ the path length, is used to evaluate the distance cost;

- $d_e = (1/n) \sum_{i=0}^{n} |d_i|$ with $d_i = \mathbf{P_0} \times \mathbf{P}/\|\mathbf{P}\|$ ($\mathbf{P_0} = \overrightarrow{p_{js}p_i}$, $\mathbf{P} = \overrightarrow{p_{js}p_{je}}$) the distance between the path point $p_i$ and its closest lane $L_j$ ($L_j$ is composed of points $p_{js}$ and $p_{je}$), is used to evaluate the distance deviation from the road network;

- $\theta_e = (1/n) \sum_{i=0}^{n} |\Delta\theta_i|$ with $\Delta\theta_i = \theta_{p_i} - \theta_{p_{js}}$ the relative angle between the path point $p_i$ and its closest lane $L_j$ ($L_j$ is composed of points $p_{js}$ and $p_{je}$), is used to evaluate the direction deviation from the road network.

TABLE I. PERFORMANCE INDEXES IN EXPERIMENT 1

|  | HybridA\*-in-SS | Dijkstra | Dijkstra-in-SS | Ours |
|---|---|---|---|---|
| $t$(s) | 11.279 | 0.405 | 0.409 | **0.009** |
| $l$(m) | 132.514 | **127.172** | 131.158 | 137.599 |
| $d_e$(m) | 0.039 | 0.625 | 0.085 | **0.028** |
| $\theta_e$(rad) | 1.997 | 2.876 | 2.004 | **0.095** |

**Experiment 1**: **both the starting and goal points are in the passage**. As can be seen in Fig.6 and Table 1, The traditional *Dijkstra* algorithm has the shortest path. Due to the neglect of the road network constraints, it has the worst performance in $d_e$ and $\theta_e$. Compared with *Dijkstra*, *Hybrid-A\*-in-SS* and *Dijkstra-in-SS* both have improved performance in $d_e$ due to the introduction of the road-network-distance-derivation penalty. However, they can not guarantee a small $\theta_e$ in nature. The path length of the proposed method is 3.8% longer than that of the *Hybrid-A\*-in-SS* mainly due to the shortcut at the goal point. However, the proposed method has better performance in the consistency with the road network, especially in terms of the direction deviation $\theta_e$, which significantly improves the navigation safety of the robot. Besides, due to the adoption of the sparse unidirectional road network in Sec. III. A, our method has better performance in planning time than those grid-map-based planners.

**Experiment 2**: **the starting and goal points are in different intersections.** Experimental results and performance indexes for Experiment 2 are shown in Fig. 7 and Table 2, respectively. Compared with other planners, our method has similar performance in path length but with shorter planning time. Due to the adoption of the two-layer-map based hybrid planning strategy in Algorithm 2, compared with *Hybrid-A\*-in-SS*, 22.86% and 91.35% improvement in $d_e$ and $\theta_e$ can be obtained, respectively. It means that with the proposed method, a much better balance between path length and the consistency with the road network has been achieved, which is vital for path planning in semi-structured environments.

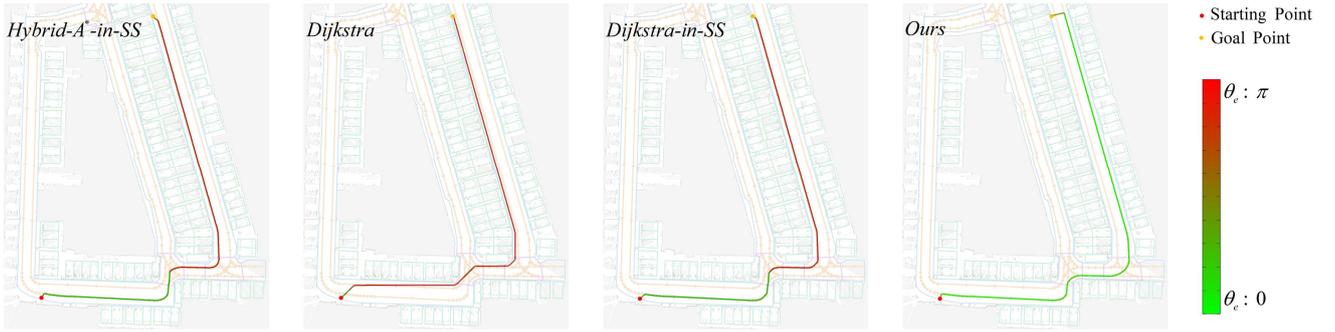

Fig. 6. Planning results in Experiment 1 with $\theta_e$ representing the angle error between the path point and its closest lane.

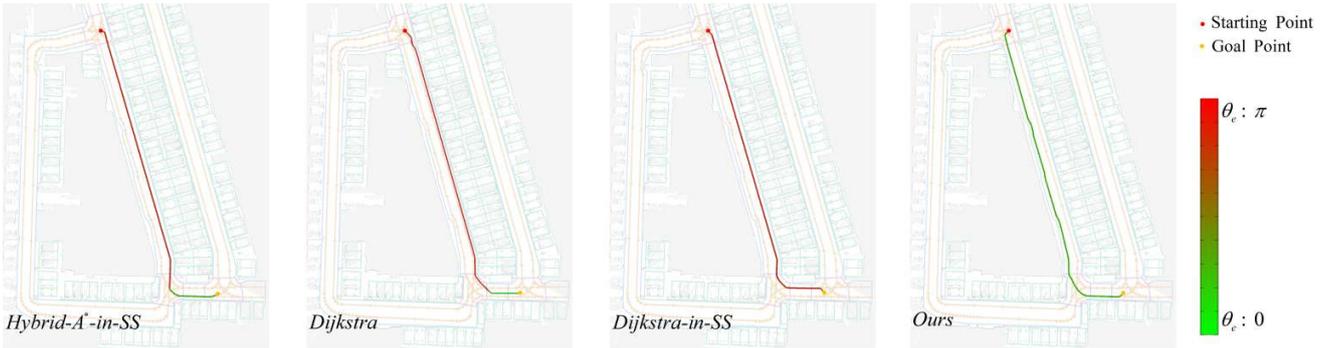

Fig. 7. Planning results in Experiment 2 with $\theta_e$ representing the angle error between the path point and its closest lane.

TABLE II. PERFORMANCE INDEXES IN EXPERIMENT 2

|  | Hybrid A*-in-SS | Dijkstra | Dijkstra-in-SS | Ours |
|---|---|---|---|---|
| $t$(s) | 8.534 | 0.349 | 0.351 | **0.101** |
| $l$(m) | 91.7328 | **88.023** | 88.479 | 91.280 |
| $d_e$(m) | 0.035 | 0.606 | 0.063 | **0.027** |
| $\theta_e$(rad) | 2.542 | 2.725 | 3.029 | **0.220** |

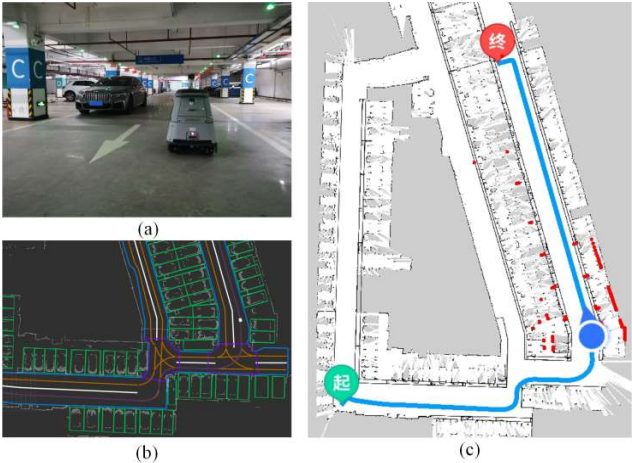

Fig. 8. An instance of the field experiment with the proposed global path planner in the garage of Guangzhou Shiyuan Electronic Technology Co., Ltd.

### B. Experiments with robots

Field experiments are carried out in the garage shown in Fig.8(a) with the semantic map in Fig.8(b). The starting and goal points in Fig.8(c) are the same as those in Experiment 1. The robot used in the experiment is a commercial garage-cleaning robot produced by Guangzhou Shiyuan Electronic Technology Co., Ltd. with RK3399 the computation unit. The global path planner proposed in this article provides a reference line to the local path planner module, which is a lightweight state lattice planner. The video of the experiment is submitted as a supplementary material.

## V. CONCLUSION

This article proposes a general and systematic global path planning method for robots in semi-structured environments. Comparative experimental results show that it achieves a much better balance between path length and the consistency with the road network, which distinguishes our work from the ones in the literature. The proposed method has been widely used in the commercial garage-cleaning robot produced by Guangzhou Shiyuan Electronic Technology Co., Ltd.

Our research focuses on solving the critical motion planning problems that prevent the commercializing robots in semi-structured environments. Research on full coverage path planning and local path planning for robots in semi-structured environments will be carried out in the future.